\documentclass{article}

\PassOptionsToPackage{numbers, compress}{natbib}
\usepackage[preprint]{neurips_2026}

\usepackage[utf8]{inputenc}
\usepackage[T1]{fontenc}
\usepackage{hyperref}
\usepackage{url}
\usepackage{booktabs}
\usepackage{amsfonts}
\usepackage{amsmath}
\usepackage{amssymb}
\usepackage{nicefrac}
\usepackage{microtype}
\usepackage{xcolor}
\usepackage{graphicx}
\usepackage{caption}
\title{Prompt Dominance and Asymmetric Verifier Costs:\\[0.2ex] {\large Empirical Ablations of GRPO at 1B Scale on GSM8K}}

\author{%
  Yi Hou \\
  University of Chinese Academy of Sciences\\
  Beijing, China \\
  \texttt{houyi25@mails.ucas.ac.cn} \\
}

\begin{document}

\maketitle

\begin{abstract}
This paper studies GRPO at 1B scale from both directions: what estimator choices do to the learning signal, and what a degraded reward signal does to what is learned. We train OLMo-2-0425-1B on GSM8K with a from-scratch implementation and measure both sides in controlled sweeps, including a verifier-quality experiment that degrades the training reward and the test-time selector identically. Four results stand out. The prompt is the first-order decision: the zero-shot prompt leaves the base model at 0.08\% (its outputs are degenerate continuations, not wrong answers), so almost no group carries a gradient, and training succeeds because the 3-shot prompt reaches 18.3\%. At this scale the estimator variants sit within seed noise, with Dr.~GRPO ahead on both seeds. In the off-policy regime, clipping is the whole story: training on data without a clipped ratio loses 4--6 points relative to the on-policy reference, while GRPO-style clipping and GSPO recover the loss entirely. Finally, the same weak verifier is far cheaper in RL than in test-time selection: a 10\%-flip verifier leaves RL's attainable gain intact (91\% and 106\% retained across two seeds) where selection retains 57\%, and a format-only verifier leaves RL with 16--30\% of its gain and selection with essentially nothing. Flip noise acts as an affine transform on the expected reward, and the group-normalized advantage with Adam's rescaling removes it exactly; the residual is a second-order variance effect that the matched-step comparison at a 30\% flip rate tests.
\end{abstract}

\section{Introduction}
\label{sec:intro}

This paper is an implementation and replication study of Group Relative Policy Optimization (GRPO) \citep{shao2024deepseekmath}, written from scratch, that improves a 1B base model on grade-school math word problems. The first half studies the estimator: a standard run, a learning-rate sweep, a prompt ablation, four one-change variants, and an off-policy correction sweep; Sections~\ref{sec:baselines}--\ref{sec:offpolicy} present those. Section~\ref{sec:verifier-costs} takes up a question that is usually only flagged in passing: \emph{the same verifier is used twice} (once as the training reward, once as a selector at test time), and these two uses are usually studied separately.

Reinforcement learning from verifiable rewards (RLVR) assumes the verifier is right. When it is not, the reward is a proxy, and proxy rewards are what get optimized \citep{amodei2016concrete,skalse2022defining,gao2023scaling}. For a large language model, the same proxy usually has a second job: deciding which of $n$ sampled answers to submit. Test-time selection and RL training degrade with verifier quality, but there is no reason for them to degrade \emph{equally}, because the mechanics differ: training averages the noisy reward over a group of samples and then takes a gradient step, while selection takes an argmax over the noisiest part of the distribution: the top of the ranking, where candidates are closest together. We test whether the asymmetry is visible at 1B scale, using a verifier whose quality we control exactly: the symbolic-equality grader used in training, with its verdicts flipped at rate $\epsilon$, or replaced by a check of the answer \emph{format} alone.

\paragraph{Code and data.} The implementation, the run records behind every number, and the scripts that render the figures are at \url{https://github.com/Helios-YQH/llm-from-scratch}.

\paragraph{Contributions.}
\begin{itemize}
  \item The prompt decides whether any learning signal exists: the reward statistics of the zero-shot and question-only arms show it directly (Section~\ref{sec:prompt}).
  \item A four-way estimator sweep and a four-way off-policy sweep, each with two seeds, reported as dot plots rather than means, so the reader can see where the seed spread is larger than the effect (Sections~\ref{sec:estimators} and~\ref{sec:offpolicy}).
  \item A comparison of RL and test-time selection under the same degraded verifiers, with selection re-ranking a fixed population of 16{,}384 generations from 512 GSM8K questions. RL is markedly more robust, with a mechanism for the asymmetry (affine invariance of the normalized advantage, plus Adam's scale invariance) and a falsifiable prediction (Section~\ref{sec:asymmetry}).
\end{itemize}

\section{Setup and protocol}
\label{sec:setup}

\paragraph{Model and data.} The policy is OLMo-2-0425-1B \citep{groeneveld2024olmo}, a 1B-parameter base model, trained on the GSM8K training split \citep{cobbe2021training} (6{,}400 questions used for training, 1{,}024 held out for internal validation) and evaluated on the 1{,}319-question test split. GRPO is implemented from scratch: response log-probabilities, the group-relative advantage, the clipped surrogate objective, the aggregation, and the training step are all our own implementation; the rollout engine is vLLM with an NCCL weight-transfer channel that pushes the updated policy into the server before every batch.

\paragraph{Compute.} All runs use one shared node with 6$\times$RTX~A6000 (48\,GB); each job holds one trainer GPU and, with the vLLM rollout backend, one rollout GPU. The training runs reported in this paper used about 100 GPU-hours.

\paragraph{The verifier.} The training reward is a symbolic GSM8K grader: it extracts the answer from a \verb|\boxed{}| or the final line and compares it to the reference with a symbolic-equality check, and it separately reports whether the response has the required \verb|</think> <answer>| format. All the evaluation reported here uses this same grader on the model's outputs, never the training reward, for the reason in Section~\ref{sec:protocol}.

\paragraph{Two decoding protocols, never mixed.} Every run is evaluated at each validation step twice: greedily (temperature 0) and by sampling (temperature 1.0). The distinction is not cosmetic. At step 0 the two differ by 21 points (greedy 0.423 vs.\ sampled 0.214); later they agree on some steps and differ on others, and the gap narrows but never closes. The prompting baselines of Section~\ref{sec:prompt} were taken under the sampling protocol, so every comparison between the trained models and the baselines is made in the sampled column; the greedy column is the lower-variance view and is what the sweeps in Sections~\ref{sec:estimators} and~\ref{sec:offpolicy} use. Mixing the two would turn a 5-point improvement into a 30-point one.

\paragraph{Budget.} The standard run is 200 rollout steps; everything else is 80, chosen because the standard run's accuracy is already close to its plateau by then on one seed and comparably far along on the other (Figure~\ref{fig:e3}). Each configuration was run with two seeds (0 and 1); where we report one seed it is stated. Group size is 8 and the rollout batch is 256 responses (32 questions) per step, so one step collects 32 groups of 8.

\paragraph{Protocol hygiene.}
\label{sec:protocol}
Three protocol requirements of this setup. (i) \emph{Validation scores the exact verifier, whatever the training reward is.} Reusing the training reward for validation made the format-only arms report their format rate (near 1.0) as accuracy for every one of 80 steps; the apparent reward-hacking curves were an artefact of the evaluation wiring. (ii) \emph{Compare within one split.} The exact arm of the weak-verifier sweep scores 0.5264 on the internal validation split and 0.4443 on the test split; we use the test split everywhere and re-evaluated every checkpoint there. (iii) \emph{Paired comparisons only.} The test split is 1{,}319 questions and all arms see the same questions, so a one-point difference is meaningful for a paired comparison; unpaired error bars at this $n$ are about $\pm1.1$ points.

\section{Baselines and the standard run}
\label{sec:baselines}

\subsection{Prompting baselines}
\label{sec:prompt}

Table~\ref{tab:e1} gives the base model's GSM8K test accuracy under three prompts. The zero-shot results are the reason the rest of this paper uses a 3-shot prompt, and they read better as a failure mode than as an accuracy level: under the zero-shot \texttt{r1\_zero} prompt the model emits the answer \emph{tags} but not answers (62.6\% of responses are format-correct, 0.08\% are correct), and a large fraction of the format-correct responses are degenerate continuations of the pretraining distribution: one repeats the question back, another emits HTML-ish fragments (\verb|mode ="5" nbsp="5"|) inside a plausible-looking tag skeleton. The model has learned the shape of the output that the prompt describes without learning the task it stands for.

\begin{table}[t]
\caption{Base-model GSM8K test accuracy by prompt (1{,}319 questions, sampled at temperature 1.0). ``Format'' is the share of responses the grader accepts as well-formed; ``length'' is the mean response length in tokens.}
\label{tab:e1}
\begin{center}
\begin{tabular}{lcccc}
\toprule
Prompt & Accuracy & Format & Formatted, wrong & Length \\
\midrule
\texttt{question\_only}     & 0.23\% & 17.4\% & 17.2\% & 273 \\
\texttt{r1\_zero} (0-shot)  & 0.08\% & 62.6\% & 62.5\% & 97 \\
\texttt{r1\_zero} (3-shot)  & \textbf{18.27\%} & 94.2\% & 75.9\% & 108 \\
\bottomrule
\end{tabular}
\end{center}
\end{table}

The arithmetic that follows from the first two rows is what fixes the experiment design: with a per-question success rate of 0.08\%, the probability that a group of 8 samples is entirely wrong is $0.994$, so essentially every group has zero reward variance and contributes no gradient. A 200-step run under that prompt would spend 6{,}400 groups to produce about 40 informative ones. The third prompt raises the per-sample rate to 18.3\%, where 80\% of groups carry signal. We therefore adopt the 3-shot version throughout, without changing anything else.

\texttt{question\_only} has a different failure profile from the two \texttt{r1\_zero} prompts: it produces the longest responses (273 tokens) and the worst format rate (17.4\%). It does not fail at arithmetic; it fails to enter answer mode at all, and continues the question freely (Section~\ref{sec:e5} returns to this arm under RL).

\subsection{The standard run}
\label{sec:e3}

\begin{figure}[t]
\caption{The standard run (200 steps, 3-shot prompt). Solid lines: greedy decoding; dashed: sampling at temperature 1.0. Both seeds are shown. The dotted line is the 3-shot prompting baseline from Table~\ref{tab:e1}.}
\label{fig:e3}
\includegraphics[width=\linewidth]{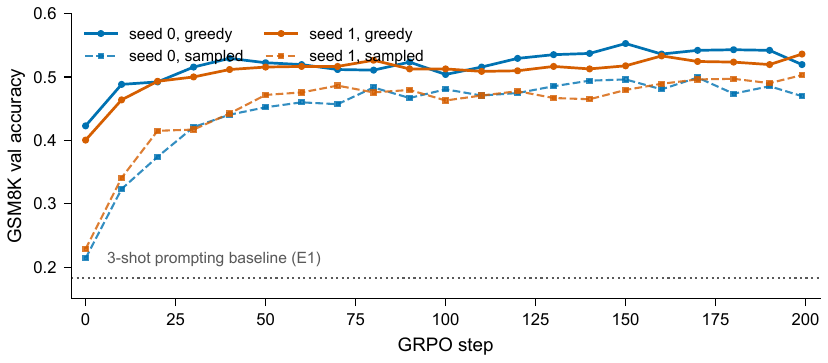}
\end{figure}

Figure~\ref{fig:e3} is the standard run, 200 steps, two seeds. Under the sampling protocol the two seeds go from 0.214 and 0.229 to 0.470 and 0.503 (the accuracy more than doubles), while the mean response length grows from 107 to 136 tokens (seed 0) and 106 to 126 (seed 1): the model learns to write longer chains, not to guess better. Format compliance decays slightly, from 0.987 to 0.931 and 0.981 to 0.933, so the gain is not bought by breaking the format.

Three secondary observations. (i) \emph{The two seeds disagree about the late curve.} Seed 0 peaks at step 150 (0.553) and ends 3.3 points below its peak; seed 1 reaches its maximum at the last step (0.536 at step 199). (ii) \emph{Entropy falls fast and then stops.} Token entropy drops from 0.72 to about 0.11 within the first 120 steps and stays in the 0.10--0.12 band for the remaining 80, low enough to be a concern but not collapse: accuracy keeps improving in that band. (iii) \emph{The fraction of the batch that carries a gradient is a moving target.} GRPO's zero-variance groups are pruned, and how many survive varies with the batch: 184 of 256 at the start for seed 0, 144 by its last step; 112 at the start for seed 1, 168 at the end. The nominal batch is 256, but the effective batch is meaningfully smaller and not fixed.

\subsection{Learning rate}
\label{sec:e4}

\begin{figure}[t]
\caption{Learning-rate sweep (80 steps, seed 0, greedy decoding). $3\times10^{-5}$ peaks at step 30 and then falls by 10 points: divergence, not slow convergence; $3\times10^{-6}$ has not converged by step 80.}
\label{fig:e4}
\includegraphics[width=\linewidth]{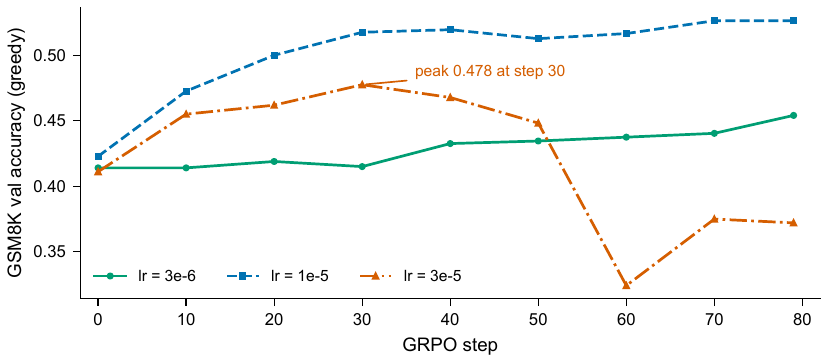}
\end{figure}

Figure~\ref{fig:e4} sweeps the learning rate over one order of magnitude, 80 steps each. $3\times10^{-5}$ is the informative one: it reaches 0.478 at step 30 and then \emph{falls} to 0.372 by step 80 (with the sampled column falling too), which is divergence rather than the tail of a plateau. $3\times10^{-6}$ is still climbing at step 80 (0.454, still its maximum at the end) and its token entropy is still 0.516, against 0.143 for the chosen rate: the entropy reveals an under-trained run before the accuracy does. The chosen rate, $1\times10^{-5}$, ends at 0.526 and is the on-policy reference used by the estimator and off-policy experiments below.

\subsection{The prompt ablation under RL}
\label{sec:e5}

\begin{figure}[t]
\caption{\texttt{question\_only} under GRPO (80 steps, both seeds, greedy decoding). The same configuration that more than doubles accuracy with a 3-shot prompt barely moves here, and the two seeds disagree by a factor of seven.}
\label{fig:e5}
\includegraphics[width=\linewidth]{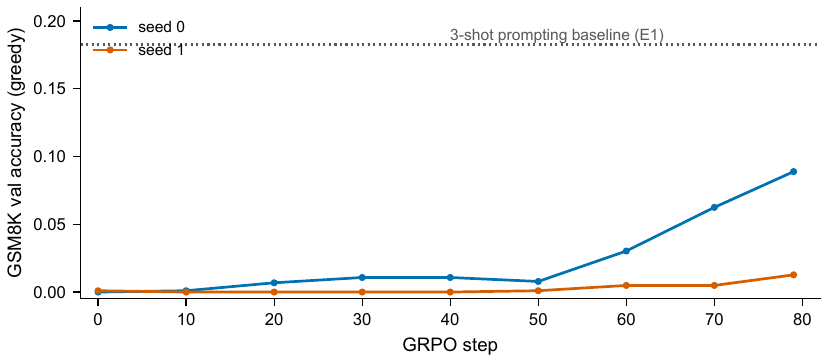}
\end{figure}

Figure~\ref{fig:e5} reports the standard configuration under the bare-question prompt, whose per-sample base rate is 0.23\%. The arm is extremely seed-dependent: seed 0 climbs from 0.000 to 0.089 (still rising at step 80) while seed 1 sits at 0.013. At the start of training each step contains, in expectation, about 0.6 correct responses out of 256, so which few responses happen to be correct dominates the gradient; the two seeds are effectively sampling different problems. Seed 0's trajectory is also the effect in miniature: RL against \emph{no examples} recovers about half of what the 3-shot prompt provides (0.089 vs.\ 0.183), by learning the answer format (the boxed-answer rate rises from 0.23 to 0.69) and then some arithmetic, from a starting point where the format does not exist. The best case observed is 0.089, still climbing at the end of its budget.

\section{Estimator variants}
\label{sec:estimators}

\begin{figure}[t]
\caption{Four one-change variants of the standard estimator, 80 steps, both seeds, final greedy accuracy. Each point is one seed; the bar is the two-seed mean. The dashed line is the standard run's own 80-step value (0.526, seed 0). Variant definitions in the text.}
\label{fig:e6}
\includegraphics[width=\linewidth]{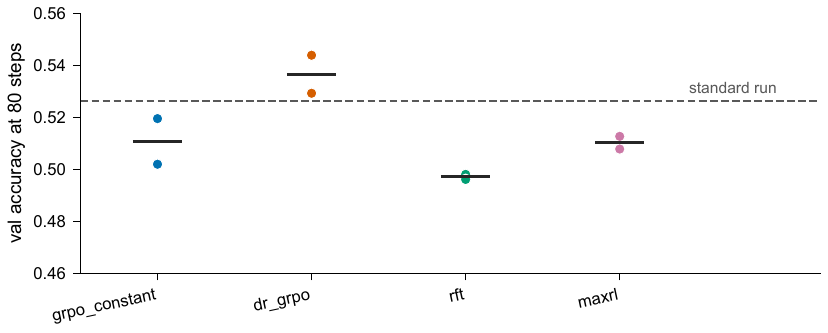}
\end{figure}

The standard estimator normalizes each sequence's loss by its length and divides the group's advantages by their standard deviation. Figure~\ref{fig:e6} changes the normalization choices one at a time: \texttt{grpo\_constant} replaces the length normalization with a constant, \texttt{dr\_grpo} additionally drops the standard-deviation division \citep{liu2025understanding}, \texttt{rft} additionally removes the group baseline so that only above-average samples are reinforced, and \texttt{maxrl} divides by the group mean instead of the standard deviation. (Each choice reweights the difficulty range of the groups: dividing by $\hat\sigma$ re-weights each group by $(\eta(1-\eta))^{-1/2}$, which up-weights the extremes (groups that are nearly all right or nearly all wrong) relative to the middle.)

The differences are small, and the figure shows why: the four variants span 0.040 in two-seed means (0.497 to 0.537), while the two seeds of a \emph{single} variant differ by 0.002 to 0.018. Only Dr.~GRPO is above the standard run on both seeds (0.529/0.544 against 0.526), by 0.003 and 0.018, the size of the seed spread itself. At this scale and horizon the normalization and baseline choices do not matter much: 80 steps at 1B parameters and 18\% accuracy is a regime where every variant is still moving. The one consistent ordering, Dr.~GRPO ahead, is also the one the derivations predict (it is the variant that removes both the length and the variance reweighting).

\section{Off-policy corrections}
\label{sec:offpolicy}

\begin{figure}[t]
\caption{Off-policy corrections (80 steps, both seeds, final greedy accuracy). Each run consumes a rollout batch as 32 sequential training steps, so the last step trains on data 32 steps stale. Only the correction scheme differs.}
\label{fig:e7}
\includegraphics[width=\linewidth]{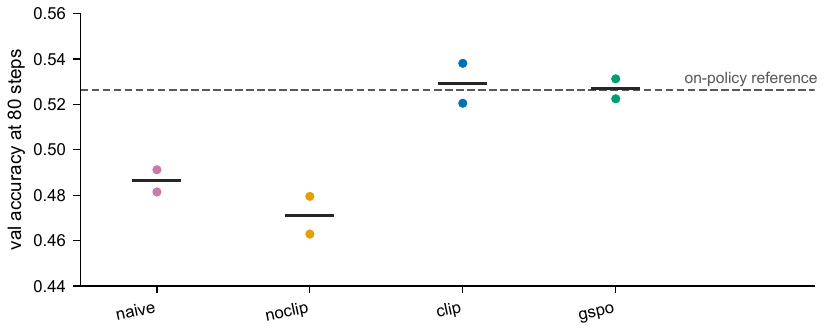}
\end{figure}

The last of the estimator experiments makes the training deliberately off-policy. The rollout batch stays 256, but the \emph{training} batch is 8 with no gradient accumulation, so the batch is consumed as 32 consecutive optimizer steps; by the end, the data is 32 policy updates stale. Four runs change only how the staleness is handled: \textsc{naive} ignores it (the objective is computed as if the data were on-policy), \textsc{noclip} multiplies each token's term by the importance ratio $\pi/\pi_{\text{old}}$ without clipping, \textsc{clip} is the standard GRPO clipped surrogate with clip range 0.2, and \textsc{gspo} uses the sequence-level ratio with a much tighter range \citep{zheng2025gspo}.

Figure~\ref{fig:e7} shows the split clearly. The two runs that use a clipped ratio sit at the on-policy reference (0.538/0.521 and 0.523/0.531 against 0.526, no measurable cost to 32$\times$ staleness), while the two that do not lose 3.5--6.3 points (0.481/0.491 and 0.480/0.463). Within that second pair, the run with \emph{unclipped} weights is the worse of the two on one seed: multiplying by an unbounded ratio amplifies the few tokens whose ratio has drifted far from 1, which is a larger error than ignoring the staleness altogether. The contrast between the clipped and unclipped columns is much larger than the seed spread, and the direction matches the mechanism: clipping is load-bearing, not decoration.

\section{What a weak verifier costs}
\label{sec:verifier-costs}

\paragraph{Design.} The verifier has two uses: it provides the training reward, and it picks the answer when several are sampled. We degrade one verifier in one way and measure both uses against the same ground truth. The degradation is applied to the symbolic grader above: \emph{flip} its verdict with probability $\epsilon$ (``noisy''), or keep only the format half of the reward (``format-only''). The sample population is fixed: 512 GSM8K questions, 32 samples each, generated once by the base model with the 3-shot prompt (16{,}384 responses). Selection re-ranks exactly that population; training uses the same verifiers as the reward and is evaluated, always, with the true grader on the full 1{,}319-question test split.

\subsection{Selection: the ceiling, and what degradation costs}
\label{sec:v3}

\begin{figure}[t]
\caption{Test-time selection over the fixed sample population (512 questions, 32 samples each). Best-of-$n$ re-ranks the same samples under verifiers of decreasing quality; ``oracle'' is the coverage ceiling (pass@$n$, the unbiased estimator of \citep{chen2021evaluating}). The horizontal line is the no-selection accuracy (n=1).}
\label{fig:v3}
\includegraphics[width=\linewidth]{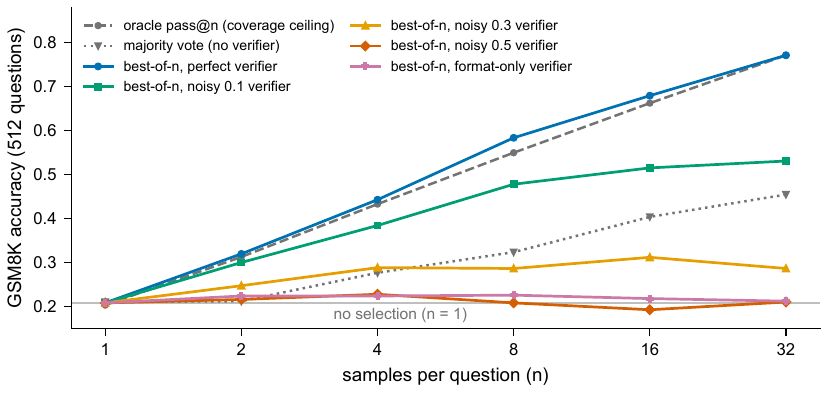}
\end{figure}

Figure~\ref{fig:v3} gives the full picture of selection. With a perfect verifier, best-of-32 reaches 0.772, which is exactly the coverage ceiling (pass@32 = 0.7715): \emph{selection can realize everything the samples contain}. Majority voting, which needs no verifier, gets 0.455, better than the base rate of 0.209 but far below the ceiling. In between, the verifier's reliability sets the height: with 10\% of verdicts flipped, best-of-32 reaches 0.531; at 30\%, 0.287; at 50\%, 0.211, statistically indistinguishable from not selecting at all. And with a format-only verifier, 0.213: a verifier that measures something real but \emph{wrong} (format correlates with correctness only weakly) is as useless as a coin flip at selecting, which is the cleanest illustration of the distinction between a verifier being informative about the response and informative about the answer.

\subsection{Training: the same verifiers as the reward}
\label{sec:v2}

\begin{figure}[t]
\caption{Training with a weak verifier (80 steps, both seeds), evaluated with the true grader on the 1{,}319-question test split. Each point is one seed, the bar is the mean; the dotted line is the untrained base model under the same protocol. The $30\%$-flip arm is absent here: it is compared on the internal validation split (Section~\ref{sec:v2}).}
\label{fig:v2}
\includegraphics[width=\linewidth]{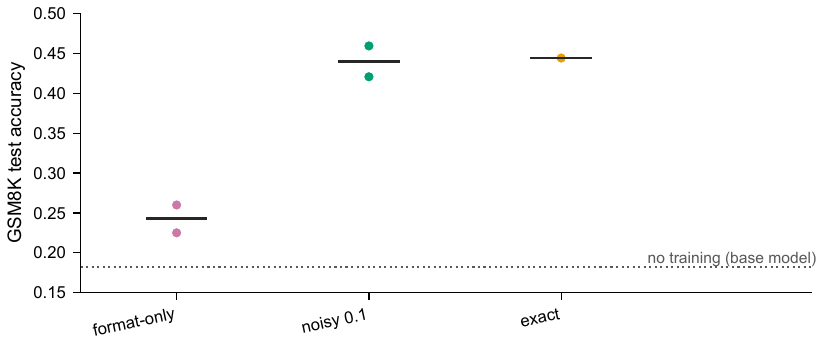}
\end{figure}

Figure~\ref{fig:v2} reports the standard 80-step configuration under three training rewards, and evaluates every arm with the true grader on the test split. The exact arm reaches 0.444, with the 3-shot prompting baseline at 0.183, so the attainable gain on this axis is 0.261. Against that yardstick: the 10\%-flip arm keeps 91\% (seed 0, 0.421) and 106\% (seed 1, 0.459), i.e.\ nothing measurable is lost to a 10\% flip rate, while the format-only arm keeps 30\% and 16\% (0.260, 0.225). The format-only arm is the pure case of proxy optimization: it reaches a format rate of 0.97, so its reward is nearly maximal, while its accuracy sits a third of the way from the base model to the exact arm. The model optimizes exactly what the reward measures; a verifier that measures the wrong thing does not slow the model down, it points it elsewhere.

A third noise level, $\epsilon=0.3$, has no test-set evaluation (an infrastructure failure interrupted it) and is compared one protocol over. On the internal validation split (1{,}024 questions, scored by the exact grader), every run has a step-60 point, before their lengths diverge: the exact arm is at 0.5166 and the 10\%-flip arm at 0.5156, while the 30\%-flip arm's two seeds sit at 0.4805 and 0.5010. Against the arms' shared starting point (0.4229) and the exact arm's gain, that is 99\% retained at $\epsilon=0.1$ and 61\% / 83\% at $\epsilon=0.3$, in the direction the prediction makes (Section~\ref{sec:asymmetry}). The two 30\%-flip seeds differ by two points, and one $\epsilon=0.1$ seed is usable on this split.

\subsection{The asymmetry, and why it should exist}
\label{sec:asymmetry}

\begin{figure}[t]
\caption{Retained fraction of the attainable gain as the verifier degrades, for the two uses of the same verifier. Selection: best-of-32 minus the base rate, over the coverage ceiling minus the base rate, on the 512-question population of Figure~\ref{fig:v3}. RL: the trained arm minus the \emph{same-protocol} prompting baseline, over the exact arm minus that baseline, on the test split; each dot is one seed. The RL series stops at $\epsilon=0.1$ because the 30\%-flip arm has no test-set evaluation; it is compared on the internal validation split in Section~\ref{sec:v2}. The dashed line is the closed form for the expected-gradient scaling of RL under flip noise, $(1-2\epsilon)$.}
\label{fig:asymmetry}
\includegraphics[width=\linewidth]{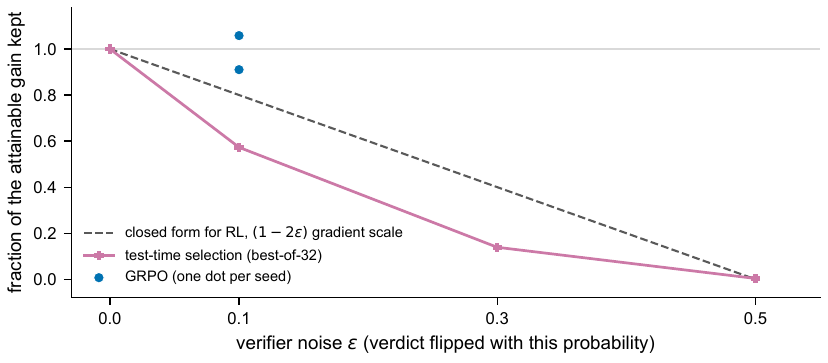}
\end{figure}

Figure~\ref{fig:asymmetry} puts the two halves on one axis. At $\epsilon=0.1$ the same noise costs selection 43\% of its attainable gain and RL nothing measurable (the two seeds land at 0.91 and 1.06 of the exact arm). The gap widens with noise: selection retains 14\% at $\epsilon=0.3$ and essentially zero at 0.5; the $\epsilon=0.3$ point for RL is compared on the internal validation split instead (Section~\ref{sec:v2}).

Two mechanisms point the same way.

\textbf{1.\ RL averages; selection takes an argmax.} A group of 8 rollouts is scored, and the update uses the group's mean reward as the baseline. Flipping one verdict moves that group's contribution by roughly $1/8$ of a reward unit, so the noise enters the gradient already averaged over 8 samples, and then over the 32 groups of a step, and then over 80 steps: an unbiased error of this kind shrinks like $1/\sqrt{\text{number of samples}}$ in the quantities that matter, and as a \emph{shift} of the reward level it disappears entirely (below). Selection has no such averaging: it keeps the single highest-scored candidate, and the candidates near the top of the ranking are the ones whose scores are closest together. At $n=32$, the gap between the best and second-best true scores is small, so a flip probability of $\epsilon$ per candidate translates into a much larger probability of choosing the wrong winner: an order statistic, and one whose sensitivity grows with $n$.

\textbf{2.\ The expected effect of symmetric flip noise is an affine transform, and GRPO is immune to affine transforms.} If the true correctness indicator is $\eta\in\{0,1\}$ and the verdict is flipped with probability $\epsilon$, then
\begin{equation}
\mathbb{E}[\tilde r \mid \eta] = (1-\epsilon)\,\eta + \epsilon\,(1-\eta) = (1-2\epsilon)\,\eta + \epsilon,
\label{eq:affine}
\end{equation}
so the expected reward is an affine function of the true one, and $\nabla_\theta\,\mathbb{E}[\tilde r] = (1-2\epsilon)\nabla_\theta\,\eta$: the noise scales the gradient and shifts the reward level, and it does not move the optimum. Now note what GRPO does with a reward vector: the advantage is $A_i = (r_i - \bar r)/\hat\sigma$, which is invariant to $r \mapsto ar + b$ for any $a>0$. The $+\epsilon$ shift cancels in the numerator and the $(1-2\epsilon)$ scale cancels in both numerator and denominator. Adam compounds this: its update is invariant to a global rescaling of the gradient as well (the first and second moments scale together). So the first-order effect of the closed form is neutralized \emph{exactly} by the estimator's normalization: what remains is the per-sample variance the flips inject, which is second order. The 10\%-flip arm is consistent with this: it does not lose the 20\% the closed form suggests, because the closed form is about the expected gradient, and the estimator normalizes exactly that away.

The second mechanism also makes a prediction, which is why the 30\%-flip arm was run: if the residual cost is the injected variance rather than the scaling, then it should grow faster than $(1-2\epsilon)$ once the flip rate is high enough for variance to dominate: an arm at $\epsilon=0.3$ should retain markedly less than the 91--106\% of the 10\% arm. The matched-step validation comparison in Section~\ref{sec:v2} is the evidence available on this: about 72\% retained across the two 30\%-flip seeds, against 99\% at $\epsilon=0.1$ on that split. The direction matches the prediction.

\paragraph{Caveats on this axis.} (i) The exact arm has a single seed, so the denominators in Figure~\ref{fig:asymmetry}'s RL numbers carry no seed variance; what is robust is that neither flip-arm seed falls far from the exact arm (0.91 and 1.06 of its gain). (ii) The selection curve and the RL points are computed on different sets (512 questions with 32 samples vs.\ 1{,}319 questions with 1 sample), so their absolute levels should not be compared, only their retained fractions. (iii) The verifier here is a program with a controlled failure mode, not a learned reward model; the numbers are properties of this protocol, and the claim we take from them is the ordering between the two uses. (iv) Format-only degrades the two uses differently but both severely, so the $\epsilon=0.1$ result is the sensitive one and it is a single noise level; the 30\% arm is the test.

\section{Related work}
\label{sec:related}

\textbf{RL with verifiable rewards.} GRPO \citep{shao2024deepseekmath} introduced the group-relative advantage used throughout this paper; the R1 recipe \citep{guo2025deepseekr1} and open replications popularized the setting. The estimator sweep tests a subset of the objective variants that have been proposed since: dropping the standard-deviation division \citep{liu2025understanding}, constant loss normalization and decoupled clipping \citep{yu2025dapo}, sequence-level importance ratios \citep{zheng2025gspo}, and rejection-sampling-style updates \citep{lambert2024tulu3}. At 1B parameters and 80 steps these choices are within seed noise of each other, with Dr.~GRPO marginally ahead on both seeds.

\textbf{Verifiers and their failure.} Training verifiers to re-rank samples is the original use of a verifier in this literature \citep{cobbe2021training}; process-level supervision is more effective but costlier \citep{lightman2023letverify,uesato2022solving}. Reward-model over-optimization (a proxy reward going up while the true objective falls) is well documented at larger scale \citep{gao2023scaling} and is the general phenomenon that the format-only arm of Figure~\ref{fig:v2} reproduces in miniature. The asymmetry measured here is a comparison between two \emph{uses} of one degraded verifier; the components (verifier quality, selection, and proxy optimization) are each well studied, and to our knowledge this paper provides the first direct comparison of the two uses under the same degradation.

\textbf{Test-time selection.} Best-of-$n$ with a verifier, and repeated sampling with a majority vote, are standard test-time-compute strategies; the coverage ceiling measured by pass@$n$ uses the unbiased estimator of \citep{chen2021evaluating}, and the observation that repeated sampling with selection converts coverage into accuracy is the ``large language monkeys'' effect \citep{brown2024monkeys}. The ceiling in Figure~\ref{fig:v3} matches the estimator exactly at $n=32$, which is the sanity check that makes the degraded-verifier curve interpretable.

\section{Limitations}
\label{sec:limitations}

Beyond the caveats stated in place: the model is 1B parameters and the task is grade-school arithmetic, so nothing here is a scaling result; every configuration has two seeds except the exact arm of the weak-verifier study, and differences below a few points are not resolved by them; the runs are 80 or 200 steps, and both the standard run and the $3\times10^{-6}$ learning rate are still improving at their last step, so these are snapshots rather than converged values; and the experiment's verifier is a program with an injected failure mode, which makes the mechanism reproducible but leaves open how much of it carries over to a learned reward model with correlated errors. One arm is evaluated on the internal validation split rather than the test split: the $\epsilon=0.3$ flip arm's re-evaluation was interrupted by an infrastructure failure, and its matched-step comparison is in Section~\ref{sec:v2}.

\section{Conclusion}
\label{sec:conclusion}

At 1B scale, on GSM8K, with an estimator implemented from scratch: the prompt is the first-order decision (0.08\% versus 18.27\% base accuracy decides whether any gradient exists at all); the reward normalization and baseline variants are second-order (within seed noise over 80 steps, with Dr.~GRPO weakly ahead); clipping in the importance-weighted surrogate is first-order when training is off-policy (4--6 points, versus none for the clipped forms); and the quality of the verifier (the thing RLVR actually optimizes) costs the two uses of that verifier very differently, with training retaining nearly all of the attainable gain under noise that destroys most of selection's. The last result has a clean mechanism in the affine invariance of the normalized advantage, and the matched-step $\epsilon=0.3$ comparison is a first test of its central prediction.

\clearpage

\appendix

\section{Reproduction details}
\label{app:repro}

\subsection{Prompts}
\label{app:prompts}

All three prompts are used verbatim; the listings wrap the long lines for
display, and the templates store the preamble as a single line. The zero-shot
and 3-shot prompts share the same preamble, and the 3-shot variant prepends
three worked GSM8K examples before the question.

\textbf{Zero-shot (\texttt{r1\_zero}).}\nobreak
{\small
\begin{verbatim}
A conversation between User and Assistant. The User asks a question, and
the Assistant solves it. The Assistant first thinks about the reasoning
process in the mind and then provides the User with the answer. The
reasoning process is enclosed within <think> </think> and answer is
enclosed within <answer> </answer> tags, respectively, i.e., <think>
reasoning process here </think> <answer> answer here </answer>.
User: {question}
Assistant: <think>
\end{verbatim}
}

\textbf{3-shot (\texttt{r1\_zero\_three\_shot\_gsm8k}).}\nobreak
{\small
\begin{verbatim}
User: There are 15 trees in the grove. Grove workers will plant trees in
the grove today. After they are done, there will be 21 trees. How many
trees did the grove workers plant today?
Assistant: <think> There are 15 trees originally. Then there were 21 trees
after some more were planted. So there must have been 21 - 15 = 6. So the
answer is 6. </think> <answer> 6 </answer>
User: If there are 3 cars in the parking lot and 2 more cars arrive, how
many cars are in the parking lot?
Assistant: <think> There are originally 3 cars. 2 more cars arrive. 3 + 2 =
5. So the answer is 5. </think> <answer> 5 </answer>
User: Leah had 32 chocolates and her sister had 42. If they ate 35, how
many pieces do they have left in total?
Assistant: <think> Originally, Leah had 32 chocolates. Her sister had 42.
So in total they had 32 + 42 = 74. After eating 35, they had 74 - 35 = 39.
So the answer is 39. </think> <answer> 39 </answer>
User: {question}
Assistant: <think>
\end{verbatim}
}

\textbf{Question-only.}\nobreak
{\small
\begin{verbatim}
{question} Please put your final answer within \boxed{}.
\end{verbatim}
}

\subsection{Hyperparameters}
\label{app:hp}

Every run shares the settings in Table~\ref{tab:hp}; the variants change only
the knobs listed in Table~\ref{tab:hpvar}.

\begin{minipage}{\linewidth}
\small
\centering
\captionof{table}{Shared training settings.}
\label{tab:hp}
\begin{tabular}{@{}ll@{}}
\toprule
Setting & Value \\
\midrule
Policy & OLMo-2-0425-1B (bf16), vLLM rollout server \\
Rollout batch & 256 responses = 32 prompts $\times$ group size 8 \\
Training batch & 256 sequences, gradient accumulation 32 \\
Optimizer & Adam, constant learning rate $10^{-5}$, grad-norm clip 1.0 \\
Loss & sequence-length normalization \\
Advantage & $(r_i - \bar r)/\hat\sigma$ (group mean, std) \\
Sampling & temperature 1.0, up to 512 new tokens \\
Evaluation & greedy and sampled (1.0) every 10 steps \\
\bottomrule
\end{tabular}
\end{minipage}

\begin{minipage}{\linewidth}
\small
\centering
\captionof{table}{Per-variant overrides (all other settings unchanged).}
\label{tab:hpvar}
\begin{tabular}{@{}lp{0.60\linewidth}@{}}
\toprule
Variant & Overrides \\
\midrule
\texttt{grpo\_constant} & constant loss normalization \\
\texttt{dr\_grpo} & constant normalization; no advantage normalization \\
\texttt{rft} & constant normalization; no advantage normalization; no baseline \\
\texttt{maxrl} & constant normalization; advantage normalized by the group mean \\
\texttt{offpolicy\_naive} & training batch 8, one step per 1/32nd of the rollout batch; no importance reweighting \\
\texttt{offpolicy\_noclip} & as naive; token-level importance ratio, unclipped \\
\texttt{offpolicy\_clip} & as naive; GRPO clipped surrogate, clip range 0.2 \\
\texttt{offpolicy\_gspo} & as naive; sequence-level ratio, range $3\times10^{-4}$ \\
degraded-verifier arms & training reward replaced by the degraded verifier (Appendix~\ref{app:verifiers}) \\
\bottomrule
\end{tabular}
\end{minipage}

\subsection{Degraded verifiers}
\label{app:verifiers}

\emph{Format-only.} The reward is 1 exactly when the response contains both
\verb|</think> <answer>| and \verb|</answer>|; the answer is never checked.

\emph{Noisy ($\epsilon$).} With probability $\epsilon$ the exact grader's
verdict is inverted (correct scored 0, incorrect scored 1); the format
component is left untouched. The flip is drawn from a SHA-256 hash of the
seed, $\epsilon$, the response, and the ground truth, not from a global RNG,
so the noisy and exact arms consume identical random streams for sampling and
exploration. The verifier is the only difference between them.

\end{document}